# A Novel Unified Lightweight Temporal-Spatial Transformer Approach for Intrusion Detection in Drone Networks

Tarun Kumar Biswas[1], Ashrafun Zannat[2], Waqas Ishtiaq[3], Md. Alamgir Hossain[4, 5, *]

[1]Department of Computer Science and Engineering, Jahangirnagar University, Dhaka, Bangladesh

[2]Department of Computer Science and Engineering, Bangladesh Army University of Science & Technology, Saidpur, Bangladesh

[3]Independent Research

[4]State University of Bangladesh, Department of Computer Science and Engineering, Dhaka, Bangladesh

[5]Skill Morph Research Lab., Skill Morph, Dhaka, Bangladesh

[1]E-Mail: tarun.pstu@gmail.com, ORCID: https://orcid.org/0009-0003-6574-4836

[2]E-Mail: spzannat@baust.edu.bd, ORCID: https://orcid.org/0000-0001-6678-0562

[3]E-mail: waqas.ishtiaq@gmail.com, ORCID: https://orcid.org/0009-0009-8903-1308

[4]E-Mail: alamgir.cse14.just@gmail.com, ORCID: https://orcid.org/0000-0001-5120-2911

**Corresponding Author:** Md. Alamgir Hossain (alamgir.cse14.just@gmail.com)

## Abstract

The growing integration of drones across commercial, industrial, and civilian domains has introduced significant cybersecurity challenges, particularly due to the susceptibility of drone networks to a wide range of cyberattacks. Existing intrusion detection mechanisms often lack the adaptability, efficiency, and generalizability needed for the dynamic and resource-constrained environments in which drones operate. This paper proposes TSLT-Net, a novel lightweight and unified Temporal-Spatial Transformer-based intrusion detection system tailored specifically for drone networks. By leveraging self-attention mechanisms, TSLT-Net effectively models both temporal patterns and spatial dependencies in network traffic, enabling accurate detection of diverse intrusion types. The framework includes a streamlined preprocessing pipeline and supports both multiclass attack classification and binary anomaly detection within a single architecture. Extensive experiments conducted on the ISOT Drone Anomaly Detection Dataset, consisting of over 2.3 million labeled records, demonstrate TSLT-Net's superior performance with 99.99% accuracy in multiclass detection and 100% in binary anomaly detection, all while maintaining a minimal memory footprint of just 0.04 MB and 9,722 trainable parameters. These results establish TSLT-Net as an effective and scalable solution for real-time drone cybersecurity, particularly suitable for deployment on edge devices in mission-critical UAV systems.

## 1. Introduction

The widespread deployment of drones across sectors such as transportation, surveillance, agriculture, and logistics has significantly expanded the attack surface for cyber threats targeting unmanned aerial systems. As drones increasingly rely on wireless networks for communication and coordination, they become highly vulnerable to a range of cyberattacks, including denial-of-service (DoS), man-in-the-middle (MITM) attacks, spoofing, payload manipulation, and replay attacks. These intrusions can severely compromise operational safety, data integrity, and mission success, posing risks not only to the drones themselves but also to surrounding infrastructure and human populations [1], [2]. Despite the growing recognition of these threats, current security mechanisms are inadequate for reliably detecting and mitigating drone-specific intrusions, leaving critical gaps in the protection of drone networks [3], [4].

Addressing the security vulnerabilities of drone networks is essential to ensure the safe and reliable integration of drones into critical operations and public environments. Without effective intrusion detection, malicious actors can exploit network weaknesses to disrupt missions, steal sensitive data, or cause physical harm, undermining trust in drone technologies and limiting their broader adoption. Developing robust and efficient detection mechanisms will not only enhance the resilience of drone systems but also enable industries and governments to deploy drones with greater confidence, unlocking the full potential of drone-based services while safeguarding public safety and infrastructure [5], [6].

Detecting intrusions in drone networks presents unique challenges due to the dynamic, resource-constrained, and heterogeneous nature of these systems. Unlike conventional networks, drones operate under strict latency, bandwidth, and energy constraints, limiting the feasibility of computationally intensive security solutions [7]. Additionally, the diversity of attack vectors, rapidly evolving threat landscapes, and the scarcity of large, labeled datasets tailored to drone environments make it difficult to train accurate and generalizable detection models. Furthermore, the need to distinguish subtle anomalies from normal drone behavior demands advanced methods capable of capturing both temporal and spatial dependencies within network traffic, which many traditional and deep learning approaches fail to address effectively [8].

The increasing adoption of drones in diverse sectors, such as logistics, surveillance, and emergency response, has significantly expanded the attack surface in drone communication networks. These systems are highly constrained in terms of latency, computational capacity, and energy, which limits the feasibility of traditional, resource-intensive security solutions. At the same time, the variety and sophistication of cyber threats targeting UAVs continue to evolve, necessitating intrusion detection systems that are not only accurate but also lightweight and scalable for real-time deployment. To address these challenges, this study introduces TSLT-Net, a novel intrusion detection framework based on a lightweight Temporal-Spatial Transformer. The main contributions of this work are summarized as follows:

- This is the first time we introduce a novel lightweight Transformer architecture that models temporal and spatial patterns in drone network traffic for intrusion detection.
- We incorporate an innovative reshaping and multi-head self-attention mechanism that captures sequential dependencies in non-sequential network traffic data, a novel adaptation in the drone security domain.
- We design an end-to-end framework capable of performing both multiclass attack classification and binary anomaly detection within a single unified model.
- We provide a scalable and generalizable solution for securing drone networks against evolving cyber threats.

Unlike existing intrusion detection methods that rely on handcrafted features, computationally intensive architectures, or models with limited generalization to drone environments [9], [10], TSLT-Net offers a lightweight yet highly expressive solution that effectively balances accuracy and efficiency. By leveraging the temporal-spatial modeling capabilities of Transformers, our approach captures complex attack signatures without compromising computational feasibility, making it suitable for real-time edge deployment. Furthermore, TSLT-Net demonstrates superior performance across both multiclass and binary detection tasks, significantly outperforming conventional machine learning and deep learning baselines, and setting a new benchmark for drone cybersecurity solutions.

The next section reviews related works and outlines current gaps in drone intrusion detection. This is followed by the proposed methodology, detailing the TSLT-Net architecture, preprocessing, and training process. The experimental results section presents performance evaluation and baseline comparisons. The paper concludes with key findings, future directions, and references.

## 2. Related Works

This section reviews recent advancements in drone intrusion detection, highlighting key contributions, limitations, and the research gaps addressed by our proposed TSLT-Net.

Intrusion detection in drone networks has recently gained growing attention due to the rising threat landscape surrounding unmanned aerial vehicles (UAVs). However, despite notable advancements in machine learning and deep learning-based IDS frameworks, several critical limitations persist, particularly in scalability, efficiency, and adaptability to resource-constrained UAV environments.

Several early studies explored traditional machine learning models for UAV intrusion detection. For instance, Al-Fuwaiers and Mishra [11] proposed an ML-based IDS, demonstrating reasonable anomaly detection performance but encountering challenges related to scalability and lack of generalization across varied drone environments. While effective in constrained settings, these approaches typically rely on handcrafted features, which limits adaptability to evolving attack patterns. To address temporal patterns in UAV traffic, Ashraf et al. [12] integrated RNN-based architectures in an IoT-enabled drone IDS. Although they reported high detection accuracy, the computational complexity and training overhead of RNNs made the solution less suitable for real-time UAV systems. Similarly, Zeng and Nait-Abdesselam [13] introduced a GAN-based IDS incorporating human-in-the-loop feedback for detecting zero-day attacks, but the added system complexity and interaction requirement hindered automation and scalability. More recent studies have applied Transformer-based and hybrid deep learning architectures. Albuali et al. [14] used transformers for forensic analysis of drone attacks, but their work was limited to post-attack analysis and did not address real-time detection. Kabir and Mowla [15] presented GIIDS, combining 1D-CNN, LSTM, TabNet, and transformers; while comprehensive, the model struggled with imbalanced datasets and incurred significant computational cost, making it suboptimal for edge devices. Vishnu and Arora [16] eveloped the READS framework using vision transformers, yet it was confined to visual modalities, neglecting network-layer traffic, which is crucial for packet-level intrusion detection.

Hybrid models such as the CNN-Transformer approach by Jia et al. [17] performed well in UAV anomaly detection but lacked flexibility across multiple intrusion types, particularly for multiclass attack scenarios. Similarly, Alotaibi et al. [18] proposed a transformer-based IDS optimized via adaptive mongoose algorithms, prioritizing energy efficiency but underperforming in detection accuracy and generalization across attack vectors.

Several other efforts targeted adjacent problem spaces. Tran et al. [19] focused on aerial video anomaly detection using transformer-based spatiotemporal models. Although effective for visual surveillance, the model does not apply to network intrusion detection. Albaseer and Khan [20] surveyed advanced IDS models for vehicle and UAV networks but did not propose specific mechanisms tailored to drone communication layers.

Hamad et al. [21] introduced XLSTM for UAV networks, showing competitive performance over some transformers, yet the model complexity and resource demands remained a barrier to real-time or edge deployment. Likewise, Alzahrani [22] explored ConvLSTM for drone security, but the model was tested only on binary anomaly detection, omitting multiclass threat detection scenarios.

Other works such as Wei et al. [23] proposed decentralized frameworks like TADAD for urban air mobility, but these remain unproven at large scale or on embedded systems. Reviews by Ozkan-Okay et al. [24] and Tlili et al. [25] emphasized the pressing need for lightweight, unified, and adaptive IDS architectures a demand that most existing models have yet to fulfill. While GAN-based approaches such as Rawat [26] (Dronedefgant) and anomaly detection from logs by Silalahi et al. [27] show promise, they are domain-specific or simulation-based, lacking real-time deployment considerations. Majumder et al. [28] introduced graph-powered detection for UAVs, a powerful approach, yet it did not explore integration with edge devices. Lastly, Andreoni et al. [29] provided a broader view of generative AI in autonomous systems, noting that although transformers hold promise, they often suffer from latency and resource inefficiencies.

Across existing studies, a clear research gap remains: no current solution simultaneously addresses the combined challenges of real-time performance, resource efficiency, and the dual task of multiclass and binary intrusion detection within a unified architecture specifically tailored for drone networks. While several approaches show promise in isolated areas, such as anomaly detection accuracy, energy optimization, or deep feature learning many are hindered by model complexity, lack of generalization, or incompatibility with edge-device constraints. As a result, these models are often heavyweight, task-specific, or not scalable for deployment in real-time, mission-critical UAV systems. In [30], They developed an intelligent federated learning algorithm that can enable multiple healthcare networks to form a collaborative security model in a personalized manner without the loss of privacy. A time-based and performance-based dynamic aggregation scheme was developed for designing a distributed training model that keeps the privacy of the CIoT devices intact to the local server [31]. In [32], they proposed a fusion strategy that was used to enhance the performance and reduce the communication overhead of detection model.

To address these limitations, we introduce TSLT-Net, a novel lightweight Temporal-Spatial Transformer framework specifically designed for efficient and accurate intrusion detection in drone communication networks. TSLT-Net uniquely combines both multiclass intrusion classification and binary anomaly detection within a single unified model, eliminating the need for task-specific IDS pipelines. The model is built upon a multi-head self-attention mechanism and a novel input reshaping strategy that enables it to capture both temporal dynamics and spatial correlations in UAV network traffic features that traditional models often overlook or process inefficiently.

## 3. Methodology

This section outlines the proposed TSLT-Net framework, detailing the dataset preprocessing, model architecture, training procedure, and evaluation strategy used for drone intrusion detection.

### 3.1 Dataset

This research employs the ISOT Drone Anomaly Detection Dataset [33], a large-scale, real-world collection of network traffic data tailored for evaluating intrusion detection systems in UAV (drone) environments. The dataset comprises approximately 2.35 million records, including both benign traffic (1.28 million samples, 54.42%) and anomalous instances (1.07 million samples, 45.58%), representing a broad spectrum of cyberattack scenarios targeting drone communication protocols.

The anomalous data spans various attack vectors such as Denial of Service (DoS) attacks, Password Cracking, Injection, IP Spoofing, Man-in-the-Middle (MITM), Replay Attacks, Video Interception, Payload Manipulation, and Unauthorized UDP Packets. Notably, DoS attacks and Password Cracking constitute over 40% of the total anomalies, indicating their prevalence in drone-targeted threat landscapes. The original PCAP-format network traces were converted into structured CSV files using modified CIC IoT scripts. Feature extraction included modifications to conventional flow metrics, integrating drone-specific attributes such as Drone_port, DS status, Entropy, and Payload Length, ensuring relevance to aerial IoT contexts.

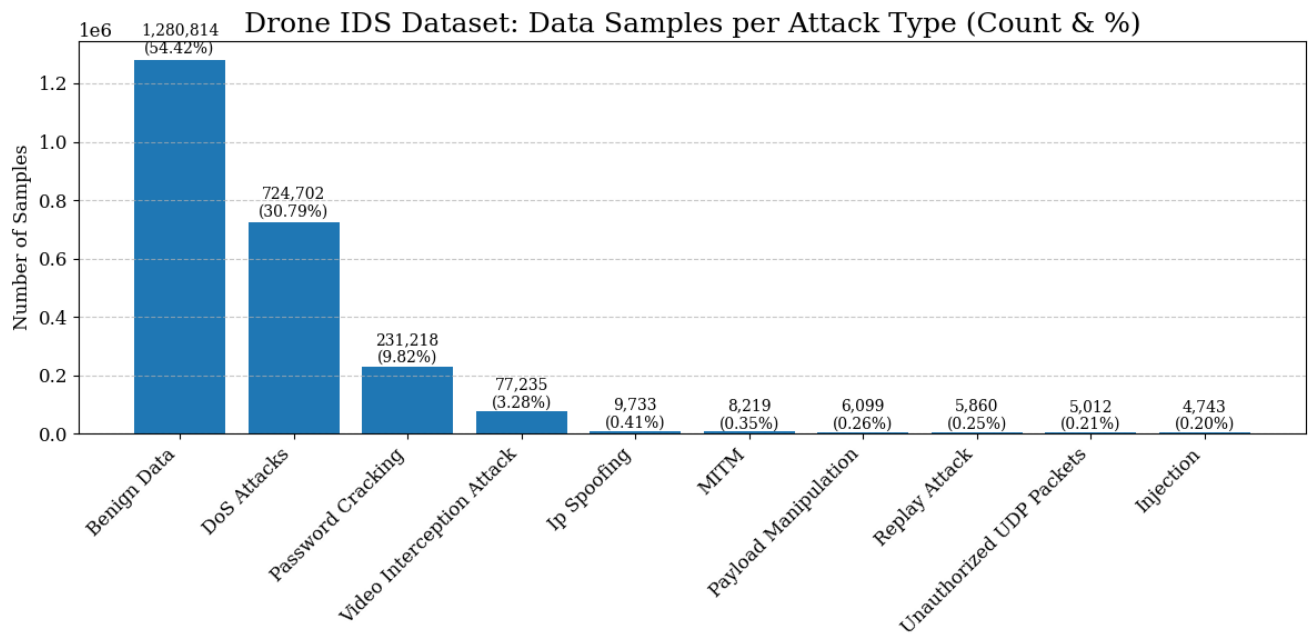

**Figure 1** Amount of Data for Each Class

As visualized in *Figure 1*, the dataset exhibits a realistic class imbalance, which is critical for evaluating detection systems under real-world conditions. Its diversity and scale provide a robust foundation for training and benchmarking machine learning and deep learning models for both multiclass intrusion classification and binary anomaly detection.

3.2 Preprocessing

The dataset underwent a structured preprocessing workflow to ensure quality and consistency. Missing values in numerical and categorical features were filled using median and mode, respectively. Numerical features were standardized using z-score normalization, while categorical features were encoded using LabelEncoder. The data was then split into input features and target labels, forming a clean, fully numeric dataset suitable for deep learning. This pipeline preserved drone-specific attributes and enabled robust training without additional sampling or feature engineering.

3.3 Model Development

This section outlines the architectural details of the baseline deep learning models, including 1D Convolutional Neural Network (CNN), Multi-layer Perceptron (MLP), Gated Recurrent Unit (GRU), Recurrent Neural Networks (RNNs), and introduces the proposed Lightweight Temporal-Spatial Transformer (TSLT-Net) designed for efficient and accurate drone intrusion detection.

*3.3.1 Model Building using 1D Convolutional Neural Network (CNN)*

In this section, the 1D Convolutional Neural Network (1D_CNN) model is designed to detect intrusions in drone networks using structured network traffic features. This 1D_CNN model can detect both benign and malicious behaviors across various cyberattack types, such as Denial-of-Service, IP Spoofing, and Payload Manipulation. Here, we train the 1D _CNN model on the ISOT Drone Anomaly Detection Dataset, and the model demonstrates strong performance with minimal computational overhead, making it suitable for real-time edge deployment.

The rise of Unmanned Aerial Vehicles (UAVs), or drones, has introduced new security vulnerabilities due to their dependency on wireless communication. These networks are increasingly targeted by cyberattacks aiming to disrupt missions or steal sensitive data. Ensuring secure drone operations, effective Intrusion Detection Systems (IDS) are required to protect this network. We proposed a deep learning-based IDS utilizing 1D CNNs, which can detect anomalies and attacks efficiently. *Figure 2* illustrates the architecture of the proposed 1D Convolutional Neural Network (CNN) model for drone intrusion detection. However, our proposed 1D_CNN model includes an input layer followed by two stacked Conv1D blocks.

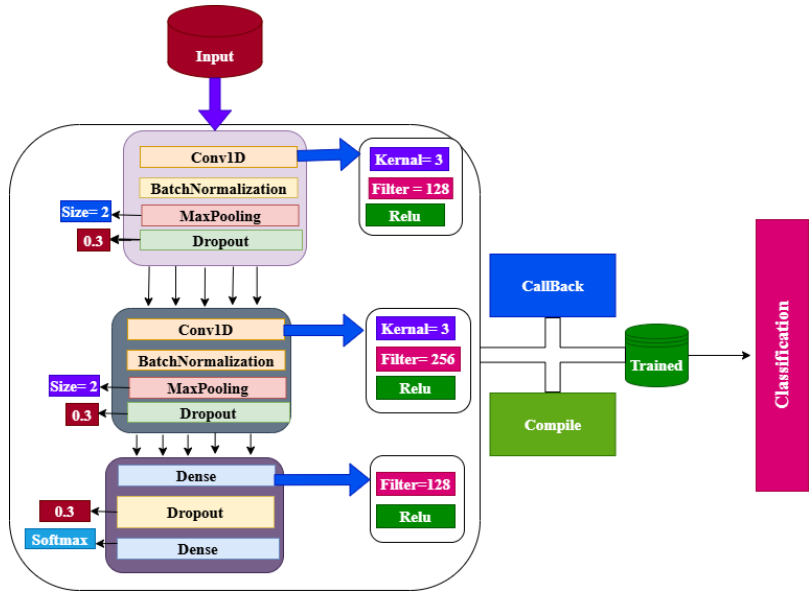

**Figure 2** Proposed 1D CNN Architecture for Drone IDS

Let X be the input data, where N is the number of samples, T is the number of time steps, and F is the number of features per time step. Let y be the labels for the samples. Let K be the number of unique classes.

Each block consists of a convolutional layer with ReLU activation (kernel size = 3), batch normalization, max pooling (pool size = 2), and a dropout layer (rate = 0.3) to mitigate overfitting. Convolutional Layer (Conv1D):

$Z_1$ = Conv1D(X, $W_1$) + $b_1$, $A_1$ = ReLU ($Z_1$) = max(0, $Z_1$), BatchNormalization: $Â_1$ = ($A_1$ - μ) / σ * γ + β, $D_1$ = Dropout(MaxPool1D($A_1$)).

Each block consists of a convolutional layer with ReLU activation (kernel size = 3), batch normalization, max pooling (pool size = 2), and a dropout layer (rate = 0.3) to mitigate overfitting. Second Convolutional Layer (Conv1D):

$Z_2$ = Conv1D($D_1$, $W_2$) + $b_2$, Activation (ReLU) and Batch Normalization: $A_2$ = ReLU($Z_2$), $Â_2$ = ($A_2$ - $μ_2$) / $σ_2$ * $γ_2$ + $β_2$

Here, we use two different sizes of filters, namely 128 and 256. Then, the extracted features are then passed to a dense layer with 128 neurons and ReLU activation, followed by another dropout layer (0.3). Global Average Pooling:

$$G = GlobalAveragePooling1D(D_2)$$

A final dense layer with softmax activation outputs class probabilities for intrusion classification. The model is compiled using the Adam optimizer and trained using callbacks (such as early stopping), enabling it to achieve optimal performance for both binary and multiclass classification tasks. Fully Connected (Dense) Layer:

$F_1$ = Dense(G, $W_3$, $b_3$) = ReLU($W_3$ * G + $b_3$), Output Layer (Dense with Softmax): ŷ = Softmax($W_4$ * $D_3$ + $b_4$)

This model is trained using early stopping (patience = 5) and a batch size of 128 over a maximum of 50 epochs. Google Colab with GPU acceleration is used for training and evaluation.

We propose an effective 1D CNN-based IDS for drone networks, capable of identifying multiple intrusion types with high accuracy. This model is computationally efficient. At this stage, we try to use this model to explore and integrating this model with transformer-based architectures like TSLT-Net for even better performance.

*3.3.2 Model Building Using Multi-layer Perceptron (MLP)*

In the second proposed deep learning architecture, we build a multi-layer perceptron (MLP) framework shown in **Figure 3**, which has three sequential blocks designed to progressively extract and refine features for classification tasks. The first block employs a dense layer with 512 filters activated by the ReLU function, followed by batch normalization and a dropout layer with a rate of 0.3 to prevent overfitting. With the same activation and regularization strategy, the second block mirrors this structure but utilizes 256 filters. The third block includes a dense layer with 128 filters, again followed by dropout, and culminates in a final dense layer with a Softmax activation function for multi-class classification. The Softmax layer outputs a probability distribution across target classes. Figure 3 shows the methodology of the multi-layer perception (MLP) framework for intrusion detection in drone network. The transformation across each block can be represented as:

$$h^l = Dropout(BatchNorm(\sigma(W^{(l)}h^{(l-1)} + b^{(l)})))$$

Where $h^l$ denotes the output of the *l-th* layer, $W^{(l)}$ and $b^{(l)}$ are the weights and biases of the dense layer, $\sigma$ represents the ReLU activation function (except for the final layer, where it is Softmax), and BatchNorm and Dropout are applied sequentially to normalize and regularize the activations.

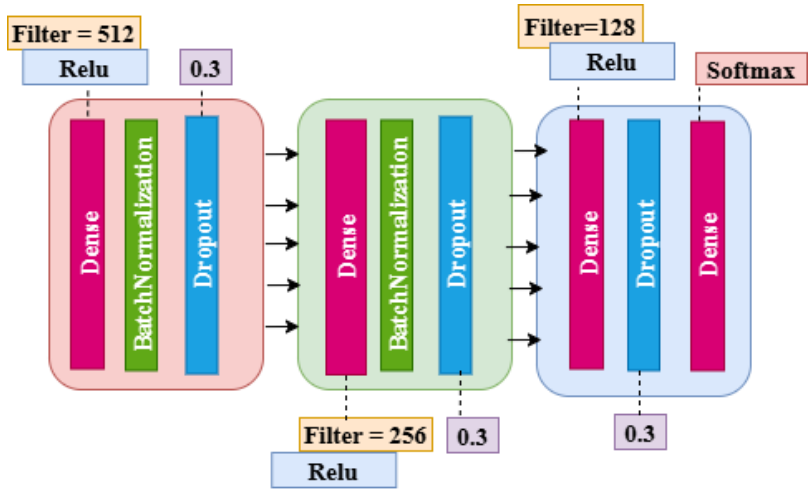

**Figure 3** proposed a multi-layer perception (MLP) framework for Drone IDS

Specifically, the hidden layers are defined as: $h^{(1)} \in R^{512}, h^{(2)} \in R^{256}, h^{(3)} \in R^{128}$. The final layer uses a Softmax activation for classification:

$$\hat{y} = Softmax(W^{(4)}h^{(3)} + b^{(4)})$$

$$\hat{y}_i = \frac{exp(z_i)}{\sum_{j=1}^{C} exp(z_j)} \text{ for } i=1,........C$$

Where C is the number of output classes and $\hat{y}_i$ is the predicted probability of class *i*.

*3.3.3 Gated Recurrent Unit (GRU) Architecture*

In this study, we create a deep learning model using Gated Recurrent Units (GRUs) for intrusion detection on drone network. For time series analysis, we can use GRU. We reshape static feature vectors into a temporal sequence format to leverage the temporal modeling capabilities of GRUs.

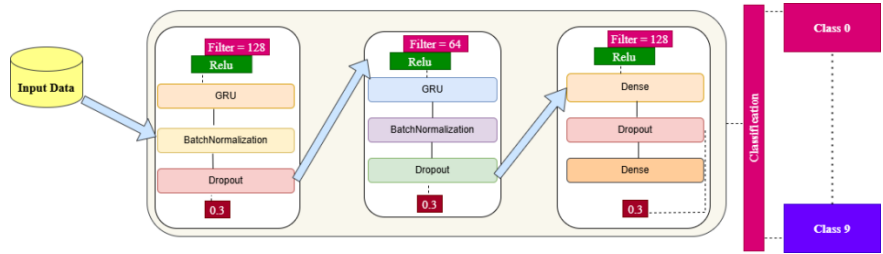

**Figure 4** proposed for Gated Recurrent Unit (GRU) Architecture for Drone IDS

*Figure 4* shows our proposed architecture integrates batch normalization, dropout regularization, and early stopping to prevent overfitting and ensure generalization. Experimental results demonstrate strong performance on the test set, indicating that GRU-based models can be adapted effectively for structured data classification. In this model, we explore the different layer for increasing our detection accuracy.

Layer 1: A GRU (Gated Recurrent Unit) layer with 128 units, outputting sequences (so the next GRU layer can process them).

$$H^1 = GRU_{128}(X), \qquad H^1 \in \mathbb{R}^{n \times 1 \times 128}$$

Layer 2: Batch Normalization normalizes activations to help with stable and faster training. Dropout randomly deactivates 30% of neurons during training to prevent overfitting.

$$H^2 = Dropout(BatchNorm(H^1)), \text{Dropout rate} = 0.3$$

Layer 3: A second GRU layer with 64 units. It does not return sequences, meaning it outputs a single vector(one per input sample).

$$H^3 = GRU_{64}(H^2), \qquad H^3 \in \mathbb{R}^{n \times 64}$$

Layer 4: Again, normalize and regularize the GRU output.

$$Z^3 = ReLu(W_3 H^3 + b_3), \; W_3 \in \mathbb{R}^{128 \times 64}, b_3 \in \mathbb{R}^{128}$$

Layer 5: Dropout applied again to prevent overfitting.

$$Z^4 = Dropout(Z^3) \qquad \text{Dropout rate} = 0.3$$

Output layer: Outputs a probability distribution over the classes using softmax.

$$Y' = \text{softmax}(W_0 Z^4 + b_0), \; W_0 \in \mathbb{R}^{C \times 128}, b_3 \in \mathbb{R}^C$$

This model is a GRU-based neural network with two recurrent layers (128 and 64 units), each followed by batch normalization and dropout (rate = 0.3), designed for sequential data classification. It ends with a dense ReLU layer and a softmax output layer for multiclass prediction.

### 3.3.4 Developing LSTM (Long Short-Term Memory) Model

In modern cybersecurity infrastructure, intrusion detection is a critical component on a drone networks. Traditional machine learning methods are not used to capture the temporal dependencies in sequential network data, limiting their effectiveness against evolving cyber threats. In this study, we present a Long Short-Term Memory (LSTM)-based deep learning model tailored for intrusion detection displayed in *Figure 5*.

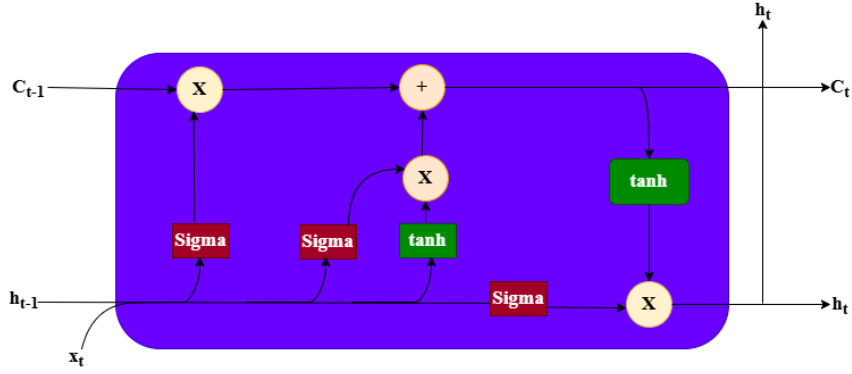

**Figure 5** LSTM (Long Short-Term Memory) Model Architecture for Drone IDS

By leveraging the temporal modeling capabilities of LSTM networks, we customize the Convolutional Neural Network layer to capture the network traffic that signal anomalous or malicious activity. This model is trained on a large-scale drone intrusion dataset that demonstrates exceptional performance. At each timestep *t*, the LSTM updates its internal cell state $C_t$, and hidden state $h_t$ using the following equations:

$$f_t = \sigma(W_f x_t + U_f h_{t-1} + b_f)$$

$$C_t = f_t \odot C_{t-1} + i_t \odot tanh(W_c x_t + U_c h_{t-1} + b_c)$$

$$h_t = o_t \odot tanh(C_t)$$

Here, $f_t$, $i_t$ and $o_t$ are the forget, input, and output gates respectively; σ denotes the sigmoid function; and ⊙ is element-wise multiplication.

*3.3.5 RNNs (Recurrent Neural Networks) Architectures*

In this research, we use two hierarchically stacked SimpleRNN layers to adopt a sequential architecture. Each layer is engineered to capture temporal dependencies within the input feature sequences. The initial recurrent layer, equipped with 128 hidden units and configured to return full sequences, facilitates deeper temporal abstraction by feeding its output into a subsequent RNN layer with 64 units.

$$X \in \mathbb{R}^{n \times d}$$

$$y \in \{0,1,\ldots,C-1\}^n$$

Where, $C$ is the number of classes. Label encoding transforms categorical labels into integers. The dataset is split into training and test sets using stratified sampling:

$$X_{train}, X_{test}, y_{train}, y_{test} = \text{train\_test\_split}(X, y)$$

Each input sample is reshaped for RNN input as:

$$x^i \in \mathbb{R}^{d \times 1}$$

so the complete training input becomes a tensor

$$X_{train} \in \mathbb{R}^{n \times d \times 1}$$

Model Architecture The model comprises multiple layers, mathematically represented as follow: RNN hidden state update at time step t: $h_t = \phi(W_{xh}x_t + W_{hh}h_{t-1} + b_h)$, where ϕ is a non-linear activation function (**ReLU**), $W_{xh}, W_{hh}$ are weights and $b_h$ is the bias. The output function, $o = \text{softmax}(W_o h_T + b_o)$

*Figure 6* represents the Batch Normalization and Dropout layers with a dropout probability of 0.3 that contribute to improved convergence and model generalization. Furthermore, our proposed model integrates a densely connected layer with 128 neurons activated via ReLU. Finally, the output layer uses a softmax activation function for multiclass classification and yields a normalized probability distribution over the target classes.

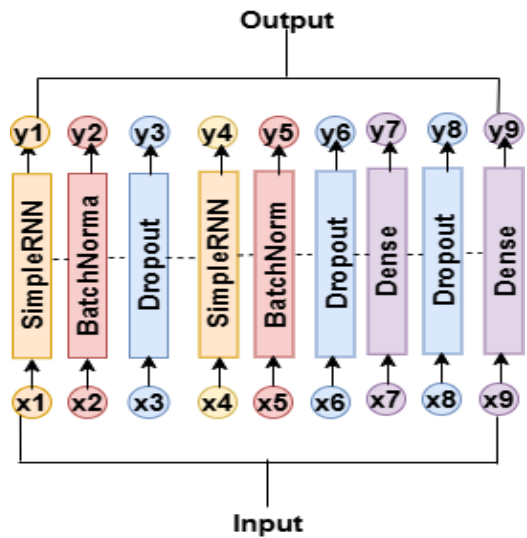

**Figure 6** Proposed Methodology for RNNs (Recurrent Neural Networks)

This architecture, combining recurrent processing with normalization and regularization techniques, is meticulously designed to capture feature interactions in our dataset, and ensure both expressive power and generalization capability during training and inference.

*3.3.6 Proposed Lightweight Temporal-Spatial Transformer (TSLT-Net)*

Our proposed lightweight Transformer-based classification model is called TSLT-Net, encompassing the stages of data preprocessing, feature encoding, model architecture design, training, and performance evaluation. *Figure 7* specifies details about each stage of the TSLT-Net model. And the detailed layer-wise configuration and parameter distribution of the proposed TSLT-Net architecture are presented in *Table 1*.

In the aspect of the commercial domain, drones have introduced significant security challenges, as drone networks become increasingly susceptible to sophisticated cyberattacks. In this research, we address the challenges of building a Lightweight Transformer-Based Classification Model (TSLT-Net), which is a sophisticated deep learning architecture. TSLT-Net is used to incorporate the core principles of the Transformer architecture, which has revolutionized the field of intrusion detection in drone technology.

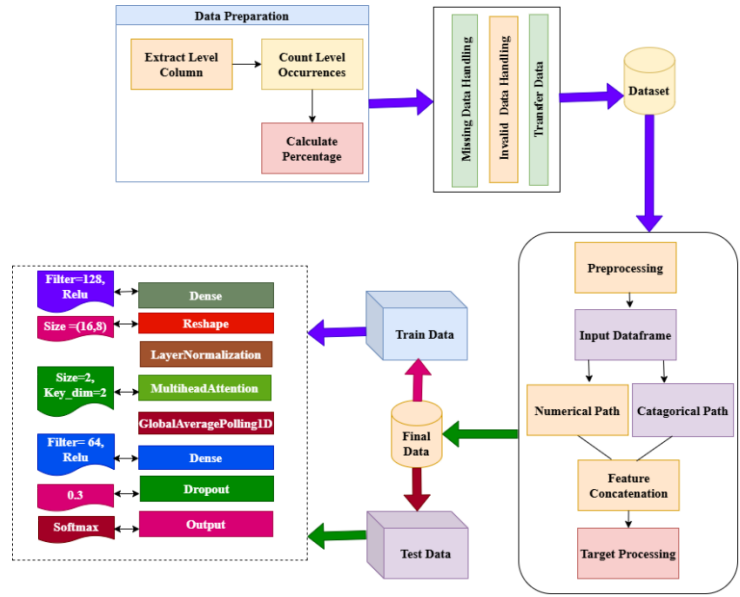

**Figure 7** Our proposed TSLT-Net Approach for Drone IDS

**Table 1** Proposed TSLT-Net Architecture – Layer-wise Configuration and Parameters

| Layer | Type | Output Shape | Activation | Parameters |
|---|---|---|---|---|
| **Input** | Input Layer | (X features,) | – | 0 |
| **Dense_1** | Fully Connected (128) | (128) | ReLU | (input_dim × 128 + 128) |
| **Reshape** | Reshape | (16, 8) | – | 0 |
| **LayerNormalization** | Normalization | (16, 8) | – | 32 |
| **MultiHeadAttention** | Self-Attention (2 heads, key_dim=4) | (16, 8) | – | 1,440 |
| **GlobalAveragePooling1D** | Pooling | (8,) | – | 0 |
| **Dense_2** | Fully Connected (64) | (64) | ReLU | 576 |
| **Dropout** | Regularization (0.3) | (64) | – | 0 |
| **Output** | Fully Connected | (num_classes,) | Softmax | 64 × num_classes + num_classes |

Our aim is to detect the intrusions in drone networks, and TSLT-Net is to combine the power of Transformer-based attention mechanisms with dense layers and pooling operations, all while keeping the model computationally efficient enough to handle large datasets without overfitting or requiring excessive resources.

Before feeding the data into TSLT-Net, we use a feature preprocessing pipeline, which is required for preparing the data. Initially, we load the dataset into a panda DataFrame, and we separate the features into categorical and numerical columns. The categorical columns represent features such as labels, IDs, or other non-numeric data, are encoded using one-hot encoding. TSLT-Net is composed of several key components, starting with an input layer x is the input vector.

A Dense layer with 128 units, followed by the activation function ReLU (Rectified Linear Unit), is used as the first step of our transformation. The activation function ReLU helps to learn the more complex relationships in the data and ensures the non-linearity of the model. Apply Dense layer with 128 units and ReLU activation. Here, the $x \in R^{d'}$ is the input vector and $W_1 \in \mathbb{R}^{128 \times d'}$ is the weight matrix maping into 128 neurons $b_1 \in \mathbb{R}^{128}$ is the base vector.

$$h_1 = ReLU(W_1 \cdot x + b_1), \text{ where } W_1 \in \mathbb{R}^{128 \times d'}$$

The reshape function is required for Transformer-based models. This reshaping results in a tensor of shape where 16 represents the sequence length, and 8 represents the feature dimension. This reshaped tensor is passed through a Layer Normalization operation, which standardizes the features to ensure that the activations do not become too large or small, thus stabilizing the learning process. The reshape equation:

$$H_2 = Reshape(h_1) \in \mathbb{R}^{16 \times 8}$$ reshape the output vector $h_1 \in \mathbb{R}^{128}$ into a 2D tensor with shape (16, 8).

Then apply a normalization layer $H_3$ = LayerNorm($H_2$). The multi-head Attention mechanism is one of the important features of Transformer architecture. In our proposed lightweight model, a multi-head attention layer has two attention heads. The attention mechanism helps the TSLT-Net to focus on different input sequences, and this ability enables the TSLT-Net to capture dependencies and relationships that might not be apparent through simple linear transformations.

Apply Multi-Head Self-Attention (2 heads, key dim = 4): $H_4$ = MHA ($H_3$, $H_3$, $H_3$), where Q = K = V = $H_3$. This equation captures the relationships across the 16-time steps using key/query/value projections of dimension 4. The multi-head attention mechanism processes the input and produces a transformed output, which is then subjected to Global Average Pooling.

$h_5 = (1/16) \times \Sigma_i H_4^{(i)}$ for i = 1 to 16, computes the Global Average Pooling by averaging all 16 time-step vectors in $H_4 \in \mathbb{R}^{128}$, resulting in a single vector $h_5 \in \mathbb{R}^8$ that summarizes the sequence features.

At this stage, the attention and pooling results apply to another Dense layer with 64 units and ReLU activation. However, this dense layer helps the model to transfer the features before passing them to the final output layer. This equation is applied for dense layer $h_6$ = ReLU($W_2 \cdot h_5 + b_2$), where $W_2 \in \mathbb{R}^{64 \times 128}$. Finally, we added a dropout layer with a rate of 0.3, which helped to reduce the risk of overfitting and improve generalization. The final transformation in the model uses the Dense output layer and this layer uses a Softmax function. The Softmax function is used for multiclass classification. The process shown in *Algorithm 1*.

**Algorithm 1** : Lightweight Transformer-Based Classification (TSLT-Net)

***Step1: Input Layer***

- *Feature matrix $X \in \mathbb{R}^{n \times d'}$*
- *label vector $y \in \mathbb{R}^n$*

***Step2: Model Building***

*1. Initialization:*

- *Initialize trainable weight matrices $W_1, W_2, W_3$ and bias vector $b_1, b_2, b_3$*
- *Set dropout rate p=0.3*

*2. Feature transformation :*

- *Compute hidden representation:*

   $h_1 \leftarrow ReLU(W_1 \cdot X + b_1)$

*3. Reshaping and Normalization:*

- *Reshape $h_1$ into $H_2 \in R^{16 \times 8}$*
- *Apply layer Normalization:*

   $H_3 \leftarrow LayerNorm(H_2)$

*4. Contextual Encoding via Attention:*

- *Apply Multi-Head Self-Attention*

   $H_4 \leftarrow MultiHeadAttention(H_3, H_3, H_3)$

*5. Dimensionality Reduction:*

- *Perform Global Average Pooling:*

   $h_5 \leftarrow GlobalAveragePooling(H_4)$

*6. Deep feature Extraction:*

- *Compute nonlinear transformation:*

  $h_6 \leftarrow ReLU(W_2 \cdot h_5 + b_2)$

*7. Regularization:*

- *Apply dropout:*

  $h_7 \leftarrow Dropout(h_6, rate=0.3)$

*8. Classification:*

- *Generate class probabilities:*

  $\hat{y} \leftarrow Softmax(W_3 \cdot h_7 + b_3)$

*9: Go to Step 3.*

*Step3: Output Layer*

- *Predicted class probabilities $\hat{y} \in \mathbb{R}^{n \times K}$*

### 3.4 Evaluation Metrics

The effectiveness of the proposed TSLT-Net with other Deep Learning Approaches was evaluated using standard metrics, including accuracy, precision, recall, and F1-score, for both multiclass and binary classification. A confusion matrix was employed to visualize class-wise prediction performance and detect any misclassifications. Additionally, model loss and accuracy curves were analyzed across training epochs to assess convergence behavior, learning stability, and generalization capability.

## 4. Experimental Results and Analysis

This section presents the experimental results and analysis of TSLT-Net, evaluating its performance on multiclass and binary drone intrusion detection tasks. We compare the proposed model against baseline methods and assess its accuracy, efficiency, and scalability.

### 4.1 Experimental Setup

All experiments were conducted on Google Colab using its GPU acceleration environment to ensure efficient model training and evaluation. The implementation was performed in Python, using TensorFlow and Keras for developing the proposed TSLT-Net and baseline deep learning models, including CNN, MLP, GRU, and RNN. Scikit-learn was employed for key tasks such as data preprocessing, train-test splitting, label encoding, one-hot encoding, standardization, and evaluation metrics including precision, recall, F1-score, and confusion matrices. Matplotlib and seaborn were used for visualizing training and validation accuracy, loss curves, and confusion matrices.

The dataset consisted of approximately 2.35 million samples, with an 80:20 split between training and testing sets, ensuring stratified sampling to preserve class balance. A grid search approach was employed for hyperparameter tuning, optimizing parameters such as learning rate, batch size, and number of layers, attention heads, and dropout rates to achieve the best balance between accuracy and efficiency. Early stopping with a patience setting of five epochs was used to prevent overfitting and ensure optimal convergence.

### 4.2 Results of the Proposed TSLT-Net Approach

*Table 2* presents the classification performance of TSLT-Net across ten multiclass categories, demonstrating exceptional precision, recall, and F1-scores for all attack types, including rare classes such as Injection, IP Spoofing, and Unauthorized UDP Packets. With an overall accuracy of 99.99%, a macro average F1-score of 0.99955, and a weighted average F1-score of 0.99998, the results highlight TSLT-Net's outstanding effectiveness and robust generalization in distinguishing diverse drone intrusion patterns.

Table 2 Classification Report of TSLT-Net for Multi-class Classification

| Label | Precision | Recall | F1-score |
|---|---|---|---|
| 0 (Benign Data) | 1.00000 | 1.00000 | 1.00000 |
| 1 (DoS Attacks) | 0.99999 | 1.00000 | 1.00000 |
| 2 (Injection) | 1.00000 | 1.00000 | 1.00000 |
| 3 (Ip Spoofing) | 0.99692 | 0.99949 | 0.99820 |
| 4 (MITM) | 1.00000 | 0.99818 | 0.99909 |
| 5 (Password Cracking) | 1.00000 | 1.00000 | 1.00000 |
| 6 (Payload Manipulation) | 1.00000 | 0.99326 | 0.99618 |
| 7 (Replay Attack) | 1.00000 | 1.00000 | 1.00000 |
| 8 (Unauthorized UDP Packets) | 1.00000 | 0.99380 | 0.99689 |
| 9 (Video Interception Attack) | 1.00000 | 1.00000 | 1.00000 |
| Accuracy | | | 0.99999 |
| Macro Avg | 0.99969 | 0.99940 | 0.99955 |
| Weighted Avg | 0.99998 | 0.99998 | 0.99998 |

*Figure 8* shows that TSLT-Net achieves near-perfect training and validation accuracy with minimal loss over epochs, indicating fast convergence, stable learning, and excellent generalization for multiclass classification.

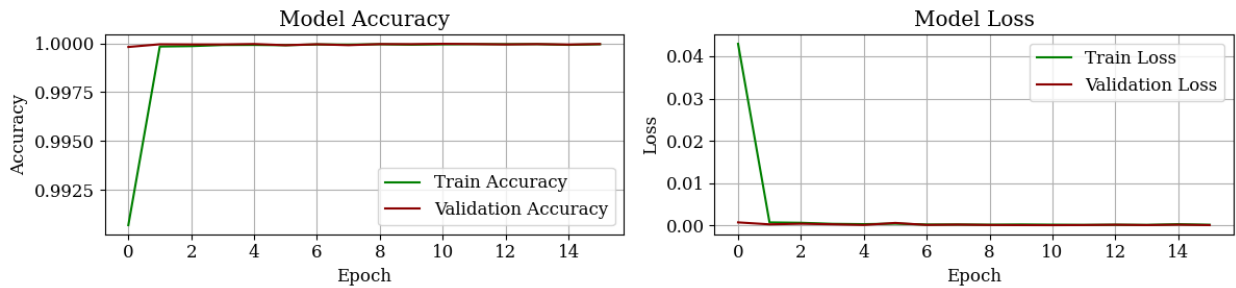

Figure 8 Accuracy and Loss for Multi-class Classification

*Figure 9* shows that TSLT-Net achieves almost perfect classification, with nearly all samples correctly predicted across all classes, confirming its excellent precision and minimal confusion even for minority attack categories.

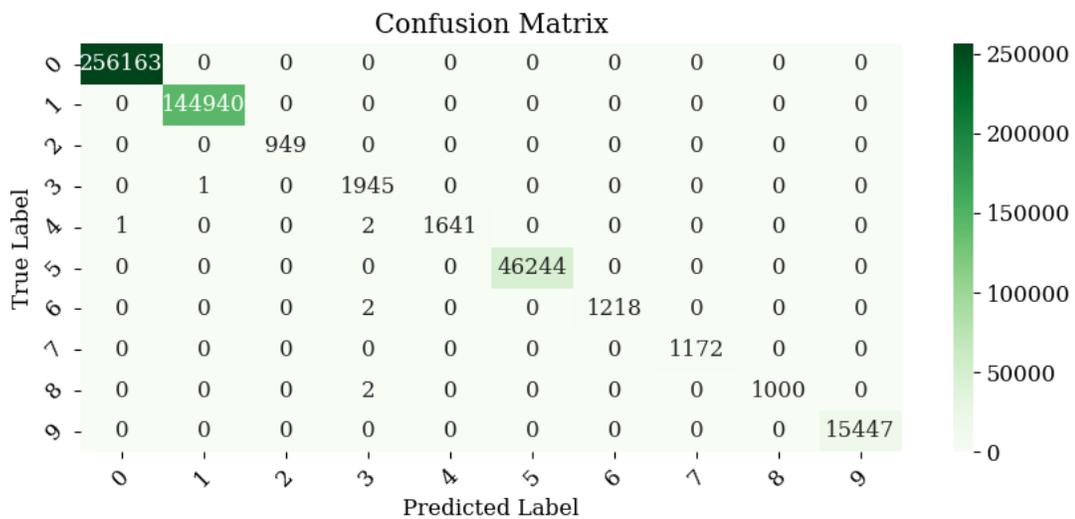

Figure 9 Confusion Matrix for Multi-class Classification

*Table 3* shows that TSLT-Net achieves perfect precision, recall, and F1-score for both benign and anomaly classes, confirming its outstanding effectiveness and reliability in binary anomaly detection tasks.

**Table 3** Classification Report for Anomaly Detection

| Label | Precision | Recall | F1-score |
|---|---|---|---|
| **Benign** | 1.00000 | 1.00000 | 1.00000 |
| **Anomaly** | 1.00000 | 1.00000 | 1.00000 |
| **Accuracy** | | | 1.00000 |
| **Macro Avg** | 1.00000 | 1.00000 | 1.00000 |
| **Weighted Avg** | 1.00000 | 1.00000 | 1.00000 |

*Figure 10* shows that TSLT-Net reaches near-perfect training and validation accuracy with minimal loss across epochs, indicating stable convergence and excellent generalization for anomaly detection tasks.

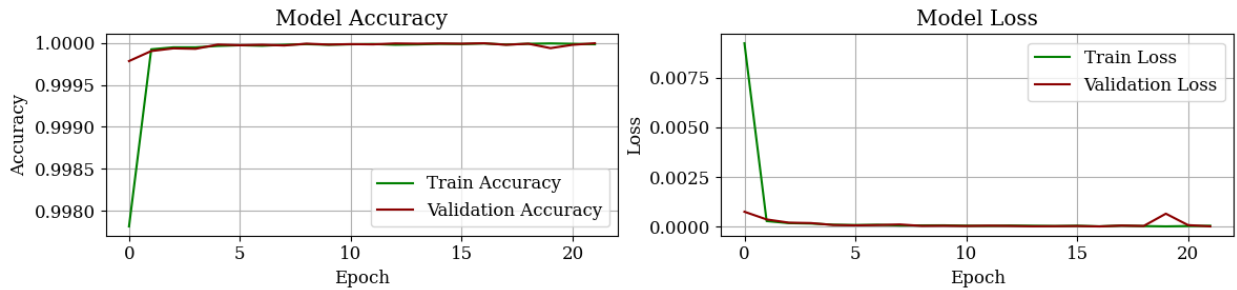

**Figure 10** Model Accuracy and Loss for Anomaly Detection

*Figure 11* shows that TSLT-Net achieves perfect classification for anomaly detection, with zero misclassifications between benign and anomaly classes, demonstrating flawless predictive performance.

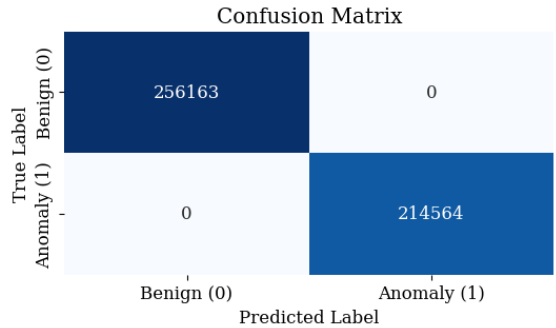

**Figure 11** Confusion Matrix for Anomaly Detection

4.3 Results for the Other Deep Learning Approaches

Figures 12 to 16 illustrate the training and validation accuracy and loss curves for the hyperparameter-tuned CNN, MLP, GRU, LSTM and RNN approaches, respectively. Across these models, notable fluctuations in validation accuracy and loss are observed, with several showing signs of instability, overfitting, or slower convergence, particularly evident in the CNN and MLP models. While GRU and RNN achieve relatively smoother training, they still display minor inconsistencies between training and validation curves, indicating less robust generalization. Compared to these baselines, the proposed TSLT-Net consistently achieves near-perfect accuracy, minimal loss, and stable learning curves, as demonstrated earlier, confirming its superior effectiveness, faster convergence, and stronger generalization capability for drone intrusion detection.

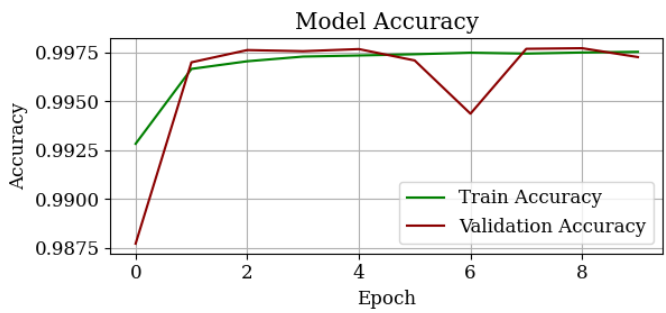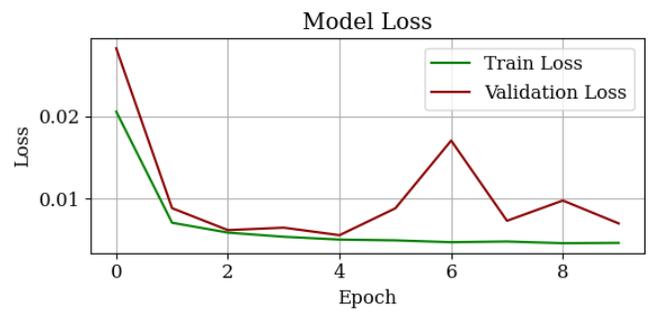

**Figure 12** Accuracy and Loss of Hyperparameter-tuned CNN Approach

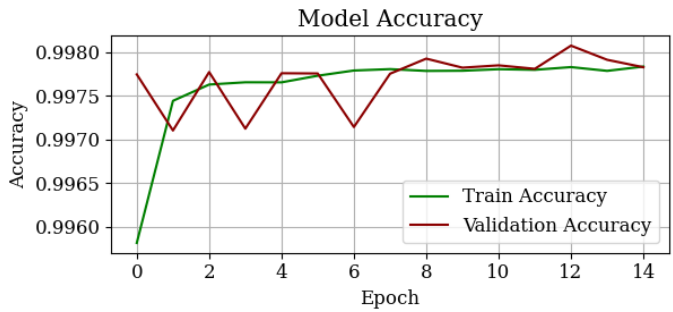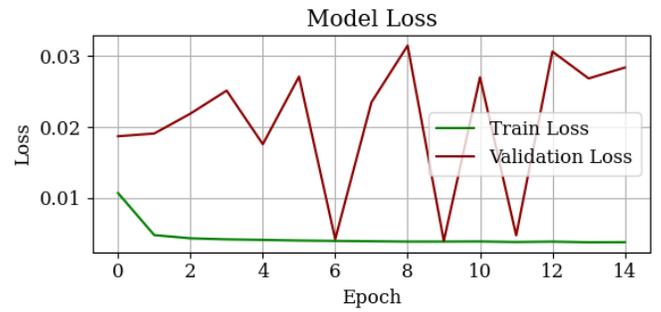

**Figure 13** Accuracy and Loss of Hyperparameter-tuned MLP Approach

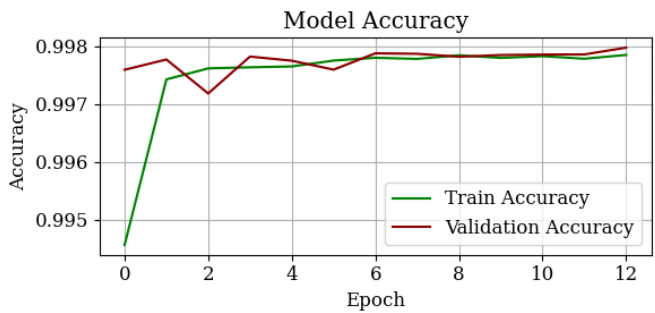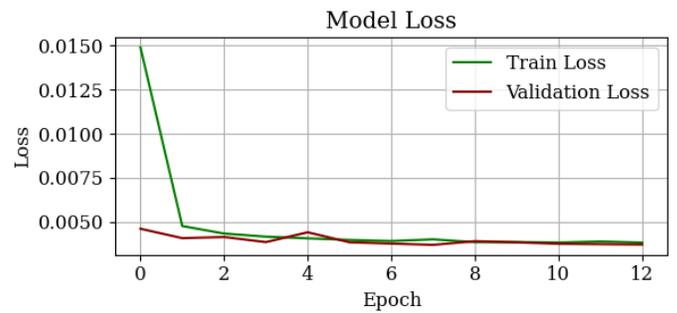

**Figure 14** Accuracy and Loss of Hyperparameter-tuned GRU Approach

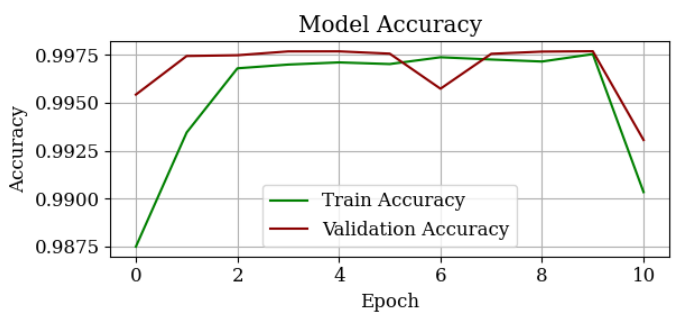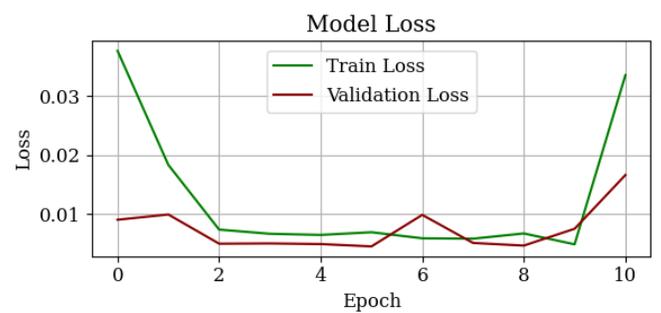

**Figure 15** Accuracy and Loss of Hyperparameter-tuned RNN Approach15

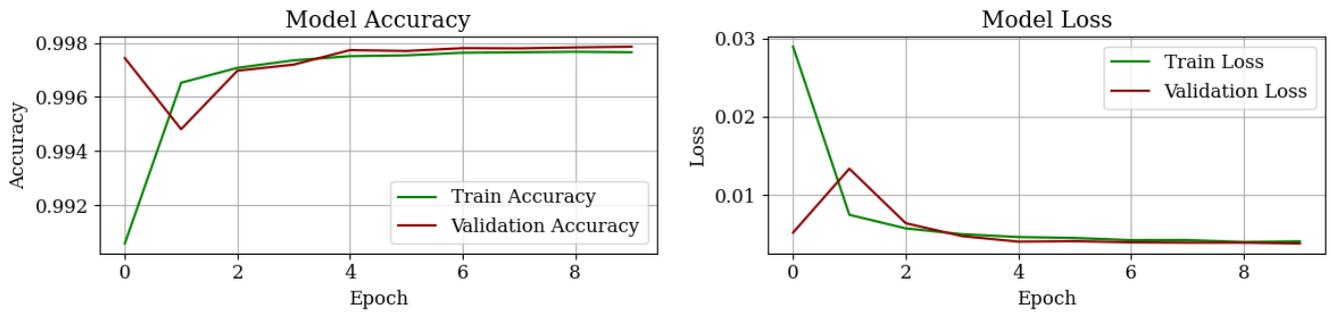

**Figure 16** Accuracy and Loss of Hyperparameter-tuned LSTM Approach

4.4 Comparison of the Newly Proposed TSLT-Net with State-of-the-art Deep Learning Models

*Table 4* presents a comprehensive comparison between the proposed TSLT-Net and several state-of-the-art deep learning models, including CNN, MLP, GRU, and RNN, across key performance metrics.

**Table 4** Comparison of the Proposed Approach with State-of-the-art Deep Learning Models

| Model | Accuracy (%) | Precision (%) | Recall (%) | F1-score (%) | Model Memory Size (MB) | Trainable Parameters |
|---|---|---|---|---|---|---|
| **CNN** | 99.78 | 99.79 | 99.79 | 99.76 | 0.51 | 134,026 |
| **MLP** | 99.72 | 99.59 | 99.72 | 99.64 | 0.76 | 199,306 |
| **GRU** | 99.79 | 99.82 | 99.79 | 99.77 | 0.46 | 120,970 |
| **RNN** | 99.78 | 99.78 | 99.78 | 99.77 | 0.15 | 38,986 |
| **LSTM** | 99.79 | 99.83 | 99.79 | 99.77 | 0.48 | 125,962 |
| **TSLT-Net (proposed)** | 99.99 | 99.99 | 99.99 | 99.99 | 0.04 | 9,722 |

*Figure 17* illustrates TSLT-Net consistently outperforms all baseline models, achieving superior accuracy (99.99%), precision (99.99%), recall (99.99%), and F1-score (99.99%), demonstrating its remarkable classification capability.

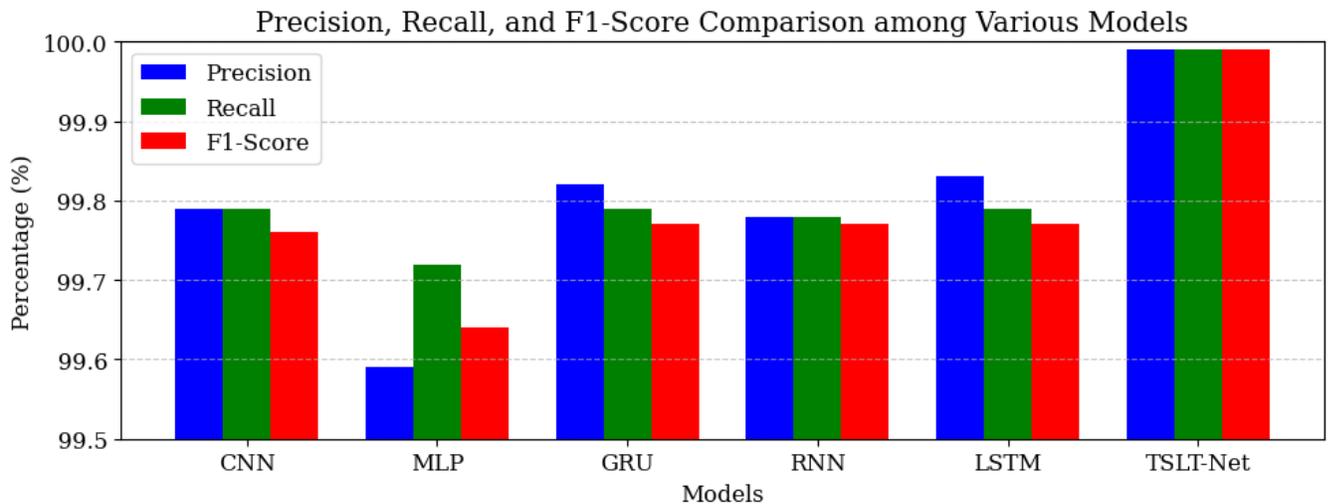

**Figure 17** Comparative Analysis of Common DL Models Proposed Approach

Notably, TSLT-Net achieves this outstanding performance with the smallest model memory footprint (0.04 MB) and the fewest trainable parameters (9,722), underscoring its exceptional efficiency for intrusion detection in drone networks. These results highlight the potential of TSLT-Net as a lightweight and high-performance solution for real-time drone intrusion detection, suitable for deployment on resource-constrained edge devices. *Figure 18* illustrates a comparative analysis of memory consumption and trainable parameters across all proposed models for intrusion classification, clearly highlighting the efficiency of each architecture in terms of computational resource requirements.

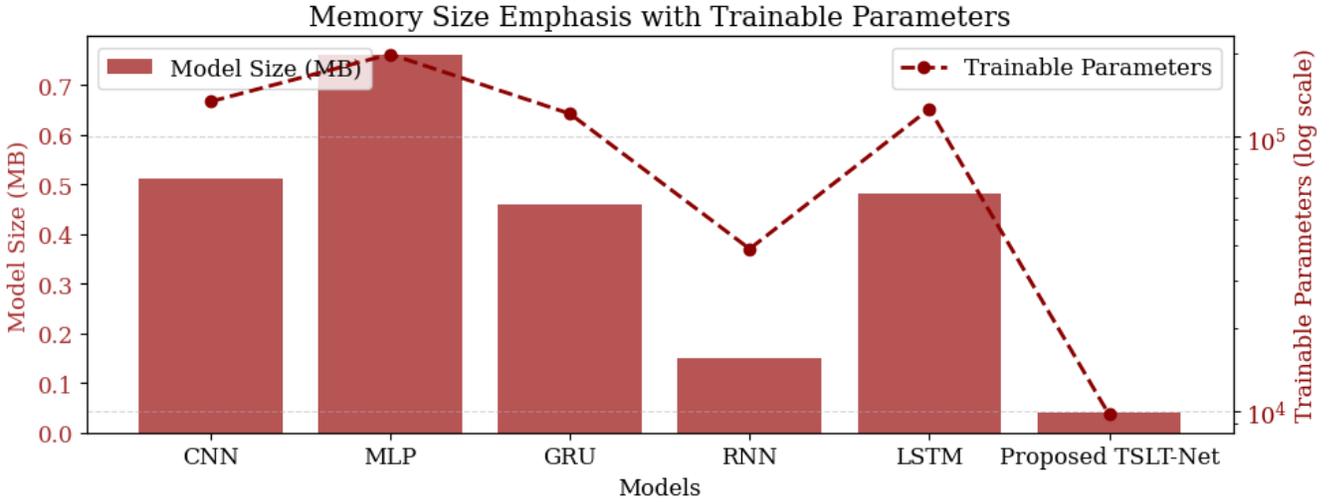

Figure 18 Comparative Analysis of Different Models size and Parameter

The proposed TSLT-Net demonstrates exceptional effectiveness by achieving near-perfect accuracy, precision, and recall in both multiclass and binary intrusion detection tasks, while maintaining a lightweight architecture suitable for real-time deployment on resource-constrained UAV systems. Table 5 presents a comparative analysis of different machine learning and deep learning approaches applied to various intrusion detection datasets.

Table 5 Comparative Analysis of state-of-art analysis.

| Reference | Model | Accuracy | Precision | Recall | F1-Score |
|---|---|---|---|---|---|
| [32] | DCNN | 0.9812 | 0.9713 | 0.978 | 0.9746 |
| [34] | FCFFN | 0.9374 | 0.9371 | 0.938 | 0.9347 |
| [35] | Ensemble | 0.996 | – | – | – |
| [36] | DT XGBOOST | 0.9085 | – | – | – |
| [37] | DAE-DNN | 0.9579 | 0.9538 | 0.958 | 0.9511 |
| [38] | Flow Transformer | 0.86 | 0.868 | – | 0.855 |
| Proposed | TSLT-Net | .9999 | 0.9999 | 0.9999 | 0.9999 |

## 5. Conclusion

This study presented TSLT-Net, a novel lightweight Temporal-Spatial Transformer architecture specifically designed for robust and efficient intrusion detection in drone networks. Addressing the rising cybersecurity threats faced by UAV systems, TSLT-Net integrates a unified detection framework capable of handling both multiclass attack classification and binary anomaly detection within a single, scalable model. The approach incorporates a streamlined preprocessing pipeline and minimal computational overhead, making it well-suited for real-time and edge-device deployment. Experimental evaluations conducted on a large-scale drone intrusion dataset demonstrated the superior effectiveness of TSLT-Net, achieving 99.99% accuracy in multiclass scenarios and 100% in binary detection, significantly outperforming established deep learning baselines including CNN, MLP, GRU, LSTM, and RNN. These results highlight the strong generalization capabilities, resource efficiency, and suitability for real-world UAV security applications of TSLT-Net. Overall, this work contributes a significant advancement toward practical, lightweight, and scalable drone intrusion detection, laying the foundation for future research in adaptive and autonomous UAV cybersecurity systems.

Future work will focus on extending TSLT-Net with federated learning and online adaptation to support real-time, privacy-preserving anomaly detection across distributed drone networks. Additionally, deploying the model on edge hardware and enhancing its robustness against adversarial attacks will further validate its practical applicability in dynamic and resource-constrained UAV environments.

**Data and Code Availability Statement**: The dataset used in this research, the ISOT Drone Anomaly Detection Dataset, is publicly available and has been clearly cited in the methodology section to ensure transparency, accessibility, and reproducibility of results. All source code developed for model implementation, training, and evaluation, including the proposed TSLT-Net architecture, is openly accessible at the following GitHub repository: https://github.com/Alamgir-JUST/TSLT-Net.

**Conflicts of interest:** There are no conflicts of interest declared by the authors.

**Funding**: No funding was received by the authors for conducting this research.

**References**

[1] N. d'Ambrosio, G. Perrone, S. P. Romano, and A. Urraro, "A cyber-resilient open architecture for drone control," *Comput. Secur.*, vol. 150, p. 104205, Mar. 2025, doi: 10.1016/j.cose.2024.104205.

[2] D. Abbadi and A. Lachkar, "The Cybersecurity Risks Threatening Drones: Innovative Solutions in the Digital Age," Jan. 09, 2025, *Social Sciences*. doi: 10.20944/preprints202501.0694.v1.

[3] S. Chiaradonna, P. Jevtić, and N. Lanchier, "Cyber risk loss distribution for various scale drone delivery systems," *Risk Sci.*, vol. 1, p. 100009, 2025, doi: 10.1016/j.risk.2024.100009.

[4] N. Constantinescu, O.-A. Ticleanu, and I. D. Hunyadi, "Securing Authentication and Detecting Malicious Entities in Drone Missions," *Drones*, vol. 8, no. 12, p. 767, Dec. 2024, doi: 10.3390/drones8120767.

[5] A. Alsumayt *et al.*, "Detecting Denial of Service Attacks (DoS) over the Internet of Drones (IoD) Based on Machine Learning," *Sci*, vol. 6, no. 3, p. 56, Sept. 2024, doi: 10.3390/sci6030056.

[6] R. Alharthi, "Enhancing unmanned aerial vehicle and smart grid communication security using a ConvLSTM model for intrusion detection," *Front. Energy Res.*, vol. 12, p. 1491332, Dec. 2024, doi: 10.3389/fenrg.2024.1491332.

[7] M. Aldossary, I. Alzamil, and J. Almutairi, "Enhanced Intrusion Detection in Drone Networks: A Cross-Layer Convolutional Attention Approach for Drone-to-Drone and Drone-to-Base Station Communications," *Drones*, vol. 9, no. 1, p. 46, Jan. 2025, doi: 10.3390/drones9010046.

[8] A. Alobaid, T. Bonny, and M. Alrahhal, "Disruptive Attacks on Artificial Neural Networks: A Systematic Review of Attack Techniques, Detection Methods, and Protection Strategies," *Intell. Syst. Appl.*, p. 200529, Apr. 2025, doi: 10.1016/j.iswa.2025.200529.

[9] Md. A. Hossain, "Deep Learning-Based Intrusion Detection for IoT Networks: A Scalable and Efficient Approach," Mar. 26, 2025, *In Review*. doi: 10.21203/rs.3.rs-6042512/v1.

[10] Y. Wu, L. Yang, L. Zhang, L. Nie, and L. Zheng, "Intrusion Detection for Unmanned Aerial Vehicles Security: A Tiny Machine Learning Model," *IEEE Internet Things J.*, vol. 11, no. 12, pp. 20970–20982, June 2024, doi: 10.1109/JIOT.2024.3360231.

[11] Department of Information Technology College of Computer and Information Sciences,Majmaah University, Majmaah, Saudi Arabia, A. Abdullah, Department of Information Technology College of Computer and Information Sciences,Majmaah University, Majmaah, Saudi Arabia, and S. Mishra, "ML-based Intrusion Detection for Drone IoT Security," *J. Cybersecurity Inf. Manag.*, vol. 14, no. 1, pp. 64–78, 2024, doi: 10.54216/JCIM.140105.

[12] S. N. Ashraf *et al.*, "IoT empowered smart cybersecurity framework for intrusion detection in internet of drones," *Sci. Rep.*, vol. 13, no. 1, p. 18422, Oct. 2023, doi: 10.1038/s41598-023-45065-8.

[13] Q. Zeng and F. Nait-Abdesselam, "Enhancing UAV Network Security: A Human-in-the-Loop and GAN-Based Approach to Intrusion Detection," *IEEE Internet Things J.*, pp. 1–1, 2025, doi: 10.1109/JIOT.2025.3545389.

[14] E. Abdulrahman Debas, A. Albuali, and M. M. Hafizur Rahman, "Forensic Examination of Drones: A Comprehensive Study of Frameworks, Challenges, and Machine Learning Applications," *IEEE Access*, vol. 12, pp. 111505–111522, 2024, doi: 10.1109/ACCESS.2024.3426028.

[15] F. Kabir, N. I. Mowla, and I. Doh, "GIIDS: Generalized Intelligent Intrusion Detection System for Heterogeneous UAVs in UAM," in *2025 27th International Conference on Advanced Communications Technology (ICACT)*, Pyeong Chang, Korea, Republic of: IEEE, Feb. 2025, pp. 377–382. doi: 10.23919/ICACT63878.2025.10936757.

[16] A. Vishnu and S. Arora, "READS: Resource efficient attack detection system for drones," *Multimed. Tools Appl.*, Oct. 2024, doi: 10.1007/s11042-024-20410-9.

[17] Y. Han, Z. Jia, S. He, Y. Zhang, and Q. Wu, "CNN+Transformer Based Anomaly Traffic Detection in UAV Networks for Emergency Rescue," 2025, *arXiv*. doi: 10.48550/ARXIV.2503.20355.

[18] S. S. Alotaibi, A. Sayed, E. Samir Abd Elhameed, O. Alghushairy, M. Assiri, and S. Saadeldeen Ibrahim, "Enhancing Security in IoT-Assisted UAV Networks Using Adaptive Mongoose Optimization Algorithm With Deep Learning," *IEEE Access*, vol. 12, pp. 63768–63776, 2024, doi: 10.1109/ACCESS.2024.3392618.

[19] D. Chauhan, H. Kagathara, H. Mewada, S. Patel, S. Kavaiya, and G. Barb, "Nation's Defense: A Comprehensive Review of Anti-Drone Systems and Strategies," *IEEE Access*, vol. 13, pp. 53476–53505, 2025, doi: 10.1109/ACCESS.2025.3550338.

[20] T. M. Tran, D. C. Bui, T. V. Nguyen, and K. Nguyen, "Transformer-Based Spatio-Temporal Unsupervised Traffic Anomaly Detection in Aerial Videos," *IEEE Trans. Circuits Syst. Video Technol.*, vol. 34, no. 9, pp. 8292–8309, Sept. 2024, doi: 10.1109/TCSVT.2024.3376399.

[21] A. H. Hamad, N. K. Hussein, and A. M. AbdulGhani, "A Deep Learning Paradigm for Intrusion Detection in Unmanned Aerial Vehicle Networks Using Extended LSTM," *Int. J. Intell. Eng. Syst.*, vol. 18, no. 4, pp. 507–523, May 2025, doi: 10.22266/ijies2025.0531.33.

[22] A. Alzahrani, "Novel Approach for Intrusion Detection Attacks on Small Drones Using ConvLSTM Model," *IEEE Access*, vol. 12, pp. 149238–149253, 2024, doi: 10.1109/ACCESS.2024.3471806.

[23] S. Wei, Z. Fan, G. Chen, E. Blasch, Y. Chen, and K. Pham, "TADAD: Trust AI-based Decentralized Anomaly Detection for Urban Air Mobility Networks at Tactical Edges," in *2024 Integrated Communications, Navigation and Surveillance Conference (ICNS)*, Herndon, VA, USA: IEEE, Apr. 2024, pp. 1–10. doi: 10.1109/ICNS60906.2024.10550825.

[24] M. Ozkan-Okay *et al.*, "A Comprehensive Survey: Evaluating the Efficiency of Artificial Intelligence and Machine Learning Techniques on Cyber Security Solutions," *IEEE Access*, vol. 12, pp. 12229–12256, 2024, doi: 10.1109/ACCESS.2024.3355547.

[25] F. Tlili, S. Ayed, and L. Chaari Fourati, "Advancing UAV security with artificial intelligence: A comprehensive survey of techniques and future directions," *Internet Things*, vol. 27, p. 101281, Oct. 2024, doi: 10.1016/j.iot.2024.101281.

[26] H. El Alami and D. B. Rawat, "DroneDefGANt: A Generative AI-Based Approach for Detecting UAS Attacks and Faults," in *ICC 2024 - IEEE International Conference on Communications*, Denver, CO, USA: IEEE, June 2024, pp. 1933–1938. doi: 10.1109/ICC51166.2024.10622524.

[27] S. Silalahi, T. Ahmad, H. Studiawan, E. Anthi, and L. Williams, "Severity-Oriented Multiclass Drone Flight Logs Anomaly Detection," *IEEE Access*, vol. 12, pp. 64252–64266, 2024, doi: 10.1109/ACCESS.2024.3396926.

[28] R. Majumder, G. Comert, D. Werth, A. Gale, M. Chowdhury, and M. S. Salek, "Graph-Powered Defense: Controller Area Network Intrusion Detection for Unmanned Aerial Vehicles," 2024, *arXiv*. doi: 10.48550/ARXIV.2412.02539.

[29] M. Andreoni, W. T. Lunardi, G. Lawton, and S. Thakkar, "Enhancing Autonomous System Security and Resilience With Generative AI: A Comprehensive Survey," *IEEE Access*, vol. 12, pp. 109470–109493, 2024, doi: 10.1109/ACCESS.2024.3439363.

[30] I. A. Khan, I. Razzak, D. Pi, U. Zia, S. Kamal, and Y. Hussain, "A Novel Collaborative SRU Network With Dynamic Behaviour Aggregation, Reduced Communication Overhead and Explainable Features," *IEEE J. Biomed. Health Inform.*, vol. 28, no. 6, pp. 3228–3235, June 2024, doi: 10.1109/JBHI.2024.3352013.

[31] I. A. Khan, D. Pi, S. Kamal, M. Alsuhaibani, and B. M. Alshammari, "Federated-Boosting: A Distributed and Dynamic Boosting-Powered Cyber-Attack Detection Scheme for Security and Privacy of Consumer IoT," *IEEE Trans. Consum. Electron.*, vol. 71, no. 2, pp. 6340–6347, May 2025, doi: 10.1109/TCE.2024.3499942.

[32] I. A. Khan *et al.*, "Fed-Inforce-Fusion: A federated reinforcement-based fusion model for security and privacy protection of IoMT networks against cyber-attacks," *Inf. Fusion*, vol. 101, p. 102002, Jan. 2024, doi: 10.1016/j.inffus.2023.102002.

[33] Z. Chen, I. Traoré, M. Mamun, and S. Saad, "Drone Anomaly Detection: Dataset and Unsupervised Machine Learning," in *Foundations and Practice of Security*, vol. 15532, K. Adi, S. Bourdeau, C. Durand, V. Viet Triem Tong, A. Dulipovici, Y. Kermarrec, and J. Garcia-Alfaro, Eds., in Lecture Notes in Computer Science, vol. 15532. , Cham: Springer Nature Switzerland, 2025, pp. 186–201. doi: 10.1007/978-3-031-87499-4_12.

[34] S. Ullah *et al.*, "A New Intrusion Detection System for the Internet of Things via Deep Convolutional Neural Network and Feature Engineering," *Sensors*, vol. 22, no. 10, p. 3607, May 2022, doi: 10.3390/s22103607.

[35] R. Alghamdi and M. Bellaiche, "An ensemble deep learning based IDS for IoT using Lambda architecture," *Cybersecurity*, vol. 6, no. 1, p. 5, Mar. 2023, doi: 10.1186/s42400-022-00133-w.

[36] S. M. Kasongo and Y. Sun, "Performance Analysis of Intrusion Detection Systems Using a Feature Selection Method on the UNSW-NB15 Dataset," *J. Big Data*, vol. 7, no. 1, p. 105, Dec. 2020, doi: 10.1186/s40537-020-00379-6.

[37] Y. N. Kunang, S. Nurmaini, D. Stiawan, and B. Y. Suprapto, "Attack classification of an intrusion detection system using deep learning and hyperparameter optimization," *J. Inf. Secur. Appl.*, vol. 58, p. 102804, May 2021, doi: 10.1016/j.jisa.2021.102804.

[38] R. Zhao *et al.*, "A Novel Traffic Classifier With Attention Mechanism for Industrial Internet of Things," *IEEE Trans. Ind. Inform.*, vol. 19, no. 11, pp. 10799–10810, Nov. 2023, doi: 10.1109/TII.2023.3241689.